# FACIAL GESTURE RECOGNITION USING CORRELATION AND MAHALANOBIS DISTANCE

**Supriya Kapoor** *(Author)*
Computer Science Engg.
Lingaya,s Institute of Mgt & Tech.
, India
supi.kap@gmail.com

**Shruti Khanna** *(Author)*
Computer Science Engg.
Lingaya,s Institute of Mgt & Tech.
,India
shruti_sagi88@yahoo.co.in

**Rahul Bhatia** *(Author)*
Information Technology Engg.
Lingaya,s Institute of Mgt & Tech.
,India
bhatia_86@yahoo.co.in

*ABSTRACT-* **Augmenting human computer interaction with automated analysis and synthesis of facial expressions is a goal towards which much research effort has been devoted recently.
Facial gesture recognition is one of the important component of natural human-machine interfaces; it may also be used in behavioural science , security systems and in clinical practice. Although humans recognise facial expressions virtually without effort or delay, reliable expression recognition by machine is still a challenge.
The face expression recognition problem is challenging because different individuals display the same expression differently.
This paper presents an overview of gesture recognition in real time using the concepts of correlation and Mahalanobis distance.We consider the six universal emotional categories namely joy, anger, fear, disgust, sadness and surprise.**

I-INTRODUCTION

The task of identifying objects and features from image data is central in many active research fields. In this paper we address the inherent problem that a single object may give rise to many possible images, depending on factors such as the lighting conditions, the pose of the object, and its location and orientation relative to the camera.

The face is the most extraordinary communicator, capable of accurately signalling emotion in a bare blink of a second, capable of concealing emotion equally well [17].

This paper presents an approach to classify different gestures. A key challenge is achieving optimal preprocessing, feature extraction and its representation, and classification, particularly under the conditions of input data variation.

From the viewpoint of automatic recognition, several various evaluation distance functions have been proposed and investigated theoretically. City block distance, Euclidean distance, weighted Euclidean distance, sub-space method, multiple similarity method, Bayes decision method and Mahalanobis distance are known typical distance functions [18]. Recognition of features in real time video is yet another challenge, due to variable characteristics such as brightness, contrast etc. which affect the video sequences or real times to a large extent. Such are difficult to analyse and work on using the earlier filter based approaches.

Results show that the Mahalanobis distance is the most effective of the seven typical evaluation distance functions. Considering the foregoing result and the properties of distribution a modified system which combines correlation and Mahalanobis distance is proposed to construct a more accurate and faster system.

The remainder of this paper is organized as follows; Section 2 briefly reviews the basics of Correlation Techniques and Mahalanobis Distance and also presents the comparison between other Distance Functions and Mahalanobis distance approach.

Section 3 gives the details of our experimental methodology.

*BACKGROUND AND RELATED WORK*

As indicated by Mehrabian [5] in face-to-face human communication only 7% of the communicative message is due to linguistic language, 38% is due to paralanguage, while 55% of it is transferred by facial expressions.

Ekman and Friesen [10] developed the most comprehensive system for synthesizing facial expressions based on what they call Action Units (AU). They defined the facial action coding system (FACS). FACS consists of 46 action units (AU), which describe basic facial movements. Traditionally, template matching methods using Eigen face by Principal Component Analysis (PCA) and Fischer face by linear discriminant analysis (LDA) are popular for face recognition and expression classification [2]. The well-known Mahalanobis Distance classifier is based on the assumption that the underlying probability distributions are Gaussian. The neural network classifiers and polynomial classifiers make no assumptions regarding underlying distributions. The decision boundaries of the polynomial classifier can be made to be arbitrarily nonlinear corresponding to the degree of the polynomial hence comparable to those of the neural networks.

Essa and Pentland [14] presented the results on recognition and singular emotional classification of facial expressions based on an optical flow method coupled with geometric, physical and motion-based face models. They used 2D motion energy and history templates that encode both, the magnitude and the direction of motion. Liu [13] used Gabor wavelet to code facial expressions. Recent studies have shown that Gabor wavelets perform better in facial expression analysis. In [1], a local Gabor filter bank which uses a part of orientation and frequency parameters is used. This reduces the dimension of feature vectors so that the computational complexity is reduced. Adaboost [3] is used as a feature selection tool on Gabor features extracted from face images and a Support Vector Machine is used to classify facial expressions.

II. ARCHITECTURAL COMPONENTS

*A.  DIGITAL IMAGE CORRELATION*

Digital Image Correlation is an optical method that employs tracking & image registration techniques for accurate 2D and 3D measurements of deformation, displacement and strain from the digital images. Thus it is important for image processing. Other applications of digital image correlation are in the field of micro- and nano-scale mechanical testing, thermo mechanical property characterization and thermo mechanical reliability in electronic packaging, stress management etc.

Correlation is a mathematical operation that is very similar to convolution. Just as with convolution, correlation uses two signals to produce a third signal. This third signal is called the **cross-correlation** of the two input signals. If a signal is correlated with itself, the resulting signal is instead called the **autocorrelation**.

The correlation between two signals (cross correlation) is a standard approach to feature detection. The amplitude of each sample in the cross-correlation signal is a measure of how much the received signal resembles the target signal, at that location. This means that a peak will occur in the cross-correlation signal for every target signal that is present in the received signal. In other words, the value of the cross-correlation is maximized when the target signal is aligned with the same features in the received signal.

For image-processing applications in which the brightness of the image and template can vary due to lighting and exposure conditions, the images can be first normalized. Normalized correlation is one of the methods used for template matching, a process used for finding incidences of a pattern or object within an image.

The peak of the cross-correlation matrix occurs where the images are best correlated.

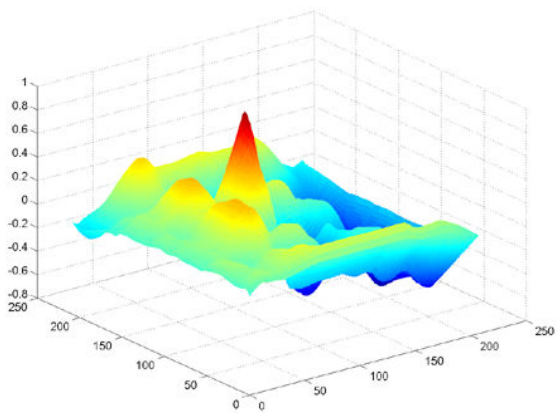

*B.  MAHALANOBIS DISTANCE*

Mahalanobis distance is a distance measure based on correlations between variables by which different patterns can be identified and analyzed. It is a useful way of determining similarity of an unknown sample set to a known one. It differs from Euclidean distance in that it takes into account the correlations of the data set and is scale-invariant, i.e. not dependent on the scale of measurements.

Distance- based approaches calculate the distance from a point to a particular point in the data set. Distance to the mean, average distance between the query point and all points in the data set, maximum distance between the query point and data set points are examples of the many options. The decision whether a data point is close to, in the data set, depends on the threshold chosen by the user [19].

The Mahalanobis distance is one of the fundamental and widely used techniques as a distance measure for classification.

According to Definition, Mahalanobis distance between two points $x = (x1.....xp)^t$ and $y = (y1......yp)^t$ in the p dimensional space $R_n$ is defined as

$$d_S(x,y) = \sqrt{(x-y)^t S^{-1} (x-y)}$$

and $d_S(x,0) = \|x\|_S = \sqrt{x^t S^{-1} x}$ is the norm of x.

Points with the same distance of the origin $\|x\|_S = c$ satisfy $x^t S^{-1} x = c^2$

which is the general equation of an ellipsoid centered at the origin and we will be interested in the distance of an observation from its center $\bar{x}$ given by

$$d_S(x, \bar{x}) = \sqrt{(x-\bar{x})^t S^{-1} (x-\bar{x})}.$$

*C.  Comparison with other distance based approaches:*

A reference model is formed for each gesture by generating a reference template (a mean vector and a covariance matrix) from the feature vector representations. Each test feature vector is compared against a reference model by distance measure or by probability estimation. Regarding the distance measure, four variations according to different usage of the covariance matrix [9] are studied. They are the City block (CBD), the Euclidean (ED), the Weighted Euclidean (WED), and the Mahalanobis (MD) distance measures.

Euclidean and Mahalanobis distance methods identify the interpolation regions assuming that the data is normally distributed (10, 11). City-block distance assumes a triangular distribution. Mahalanobis distance is unique because it automatically takes into account the correlation between descriptor axes through a covariance matrix. Other approaches require the additional step of PC rotation to correct for correlated axes. City block distance is particularly useful for the discrete type of descriptors.

Of the four base distance measures, there appears to be a significant improvement with Mahalanobis distance.

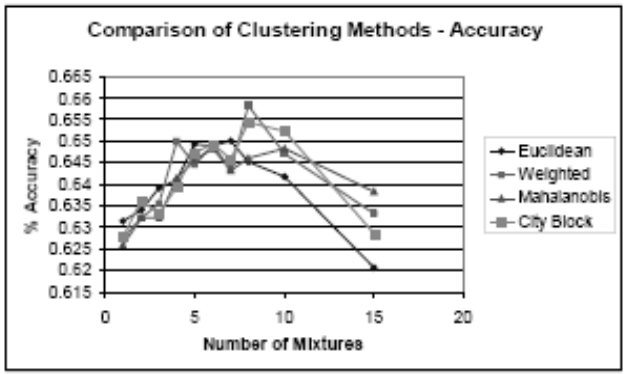

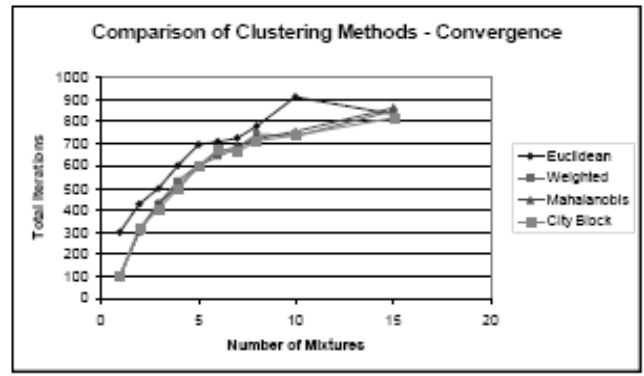

(a) Accuracy Results  (b) Convergence Results

Figure 4: Comparison of Clustering Methods

As can be seen in the Figure 4[11], there does not seem to be a great difference between the four different methods, although the Weighted Euclidean method does outperform the Euclidean method by around 1%. This is also reflected slightly in the convergence investigation, where the normal Euclidean method takes more iterations to converge. It is also interesting to note that any measure may have been used in the system, including the less computationally complex City Block distance measure.

It was found, however, that the normal Euclidean distance measure performed the most poorly and took the largest number of iterations to converge.

*D.     Disadvantages of Mahalanobis Distance:*

The drawback of the Mahalanobis distance is the equal adding up of the variance normalized squared distances of the features. In the case of noise free signals this leads to the best possible performance. But if the feature is distorted by noise, due to the squaring of the distances, a single feature can have such a high value that it covers the information provided by the other features and leads to a misclassification. Therefore, to find classification procedures which are more robust to noise we have to find a distance measure which gives less weight to the noisy features and more weight to the clean features. This can be reached by comparing the different input features to decide which feature should be given less weight or being excluded and which feature should have more weight[8].

## III. METHODOLOGY

The algorithm is as follows,

Firstly, the train images are utilized to create a low dimensional face space. This is done by performing Principal Component Analysis (PCA) in the training image set and taking the principal components (i.e. Eigen vectors with greater Eigen values). In this process, projected versions of all the train images are also created. Secondly the 2-dimensional cross correlation will be done between the Video Sequence and the Image which consists of only the part of the face and expressions that are to be correlated in the sequence. The test image obtained by the correlation is projected on the face space as a result, all the test images are represented in terms of the selected principal components. Thirdly, the Mahalanobis of a projected test image from all the projected train images are calculated and the minimum value is chosen in order to find out the train image which is most similar to the test image. The test image is assumed to fall in the same class that the closest train image belongs to. Fourthly, in order to determine the intensity of a particular expression, its Mahalanobis distance from the mean of the projected neutral images is calculated. The more the distance - according to the assumption - the far it is from the neutral expression. As a result, it can be recognized as a stronger expression.

*A.     EXTRACTION*

We say that two random variables (RVs) are correlated if knowing something about one tells something about the other RV. There are degrees of correlation and correlation can be positive or negative. The role of correlation for image recognition is not much different in that it tries to capture how similar or different a test object is from training objects. However, straightforward correlation works well only when the test object matches well with the training set.

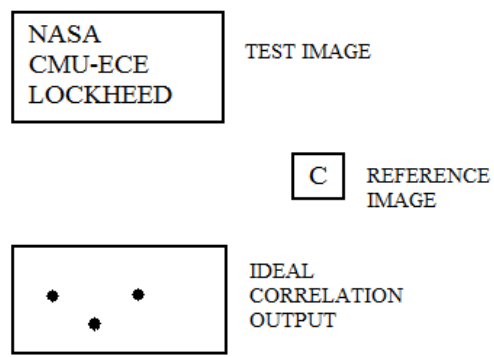

Let the reference image be represented as r [m, n]
Let the test image be represented as t [m, n]
In figure1, there are two images: a reference image of the pattern which is to be found out and a test image that
Figure1. Schematic of the image correlation: reference image, test image, and ideal correlation output contains many patterns. In this example, the letter ''C'' has to be searched.  The reference may be a client's face image stored on a smart card, and the test image may be the one he is presenting live to a camera. For the particular case in Figure 1, let us assume that the images are binary with black regions taking on the value 1 and white regions taking on the value 0. The correlation of the reference image r[m, n] and the test image t[m, n] proceeds as follows. Imagine overlaying the smaller reference image on top of the upper left corner portion of the test image. The two images are multiplied (pixel-wise) and the values in the resulting product array are summed to obtain the correlation value of the reference image with the test image for that relative location between the two. This calculation of correlation values is then repeated by shifting the reference image to all possible centering of the reference image with respect to the test image. As indicated in the idealized correlation output in Figure 1, large correlation values should be obtained at the three locations where the reference matches the test image. Thus, we can locate the targets of interest by examining the correlation output for peaks and determining if those correlation peaks are sufficiently large to indicate the presence of a reference object.

The cross-correlation of two complex functions $f(t)$ and $g(t)$ of a real variable $t$, denoted $f \star g$ is defined by

$$f \star g \equiv \overline{f}(-t) * g(t),$$

where $*$ denotes convolution and $\overline{f}(t)$ is the complex conjugate of $f(t)$.

### B. REPRESENTATION

The Facial gestures extracted by the correlation approach can be represented for the recognition task by various techniques.

It has been observed that the PCA based representation is used when distance vector measure techniques are used for classification.

In order to make the recognition task tractable, the pixel-based appearance needs to be represented by a compact coding. For this purpose, statistical redundancy reduction principles are used. Unsupervised learning techniques such as principal component analysis (PCA), independent component analysis (ICA), kernel principal component analysis, local feature analysis, and probability density estimation, as well as supervised learning techniques such as multi-linear analysis, linear discriminant analysis (LDA) and kernel discriminant analysis (KDA) exist. As for statistical unsupervised techniques, PCA can be computed as an optimal compression scheme that minimizes the mean squared error between an image and its reconstruction. This easy to compute, unsupervised, learning technique is mainly used for dimension reduction and produces uncorrelated components [20, 16]. Another representation based on multiple low-dimensional Eigenspaces is proposed in [15].

Principal Component Analysis (PCA) involves a mathematical procedure that transforms a number of possibly correlated variables into a smaller number of uncorrelated variables called principal components.

PCA is the simplest of the true eigenvector-based multivariate analyses. Often, its operation can be thought of as revealing the internal structure of the data in a way which best explains the variance in the data.

### C. RECOGNITION

The Mahalanobis distance is a very useful way of determining the "similarity" of a set of values from an "unknown: sample to a set of values measured from a collection of "known" samples.

One of the main reasons the Mahalanobis distance method is used is that it is very sensitive to inter-variable changes in the training data. In addition, since the Mahalanobis distance is measured in terms of standard deviations from the mean of the training samples, the reported matching values give a statistical measure of how well the spectrum of the unknown sample matches (or does not match) the original training spectra.

Mahalanobis, distance (Johnson and Wichern, 1998) from $\mathbf{x}$ to $\mu$, can be written

$$\Delta_{ik}^2 = (\mathbf{x}_i - \boldsymbol{\mu}_k)^T \boldsymbol{\Sigma}^{-1} (\mathbf{x}_i - \boldsymbol{\mu}_k),$$

Where $\mu_k$ are the mean and $x_i$ is the input vector of attributes where $\Sigma$ is the covariance matrix given by

$$\boldsymbol{\Sigma} = \begin{bmatrix} \sigma_{11} & \sigma_{12} & \cdots & \sigma_{1L} \\ \sigma_{21} & \sigma_{22} & \cdots & \sigma_{2L} \\ \vdots & \vdots & \ddots & \vdots \\ \sigma_{L1} & \sigma_{L2} & \cdots & \sigma_{LL} \end{bmatrix}$$

and the individual covariance values of $\Sigma$ are computed from the outer product sum given by

$$\boldsymbol{\Sigma} = \frac{1}{N} \sum_{i=1}^{N} (\mathbf{x}_i - \boldsymbol{\mu}_j)(\mathbf{x}_i - \boldsymbol{\mu}_j)^T$$

Thus, Mahalanobis distance can be seen as the generalization of Euclidean distance, and can be computed for each cluster if the covariances of the cluster are known [12].

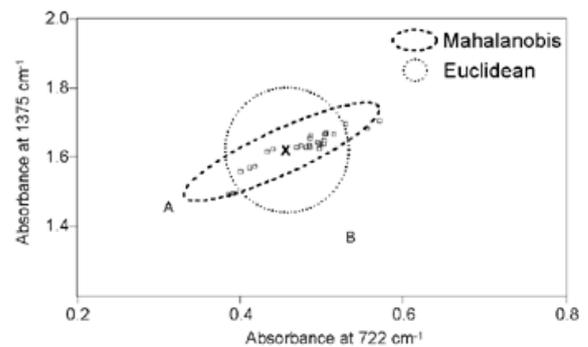

Figure 2 Absorbance of two selected wavelengths plotted against each other

An example Euclidean boundary has been superimposed on the group points in Figure 2. In addition, two hypothetical unknown sample points "A" and "B" have been added as well. Notice that although the training set group points tend to form an elliptical shape, the Euclidean distance describes a circular boundary around the mean point. By the Euclidean distance method, sample "B" is just as likely to be classified as belonging to the group as sample "A." However, sample "A" clearly lies along the elongated axis of the group points, indicating that the selected wavelengths in the spectrum are behaving much more like the training group than those same wavelengths in the spectrum of sample "B." Clearly, the Euclidean distance method does not take into account the variability of the values in all dimensions, and is therefore not an optimum discriminant analysis algorithm for this case.

The Mahalanobis distance, however, does take the sample variability into account. Instead of treating all values equally when calculating the distance from the mean point, it weights the differences by the range of variability in the direction of the sample point.

## IV. RECOGNITION EXPERIMENTS

To assess the viability of this approach to gesture recognition, we have performed experiments on real time video and built a system to locate and recognize expressions in a dynamic environment. We first collected face images under a wide range of expressions and conditions.

In the training set, there are 50 images consisting of 5 different expressions namely happy, sad, disgust, anger & neutral of size 250 X 250.

The experiments show an increase in performance accuracy as the number of images in the training set increases.

The results also indicate that changing lighting conditions causes errors while performance drops dramatically with size change.

In Real Time, people are constantly moving. Even while sitting, we fidget and adjust our body position, blink, look around and such. For the case of moving person in a static environment, we build simple motion detection and tracking system, which locates and tracks the position of head

FACIAL GESTURE RECOGNITION PROTOTYPE

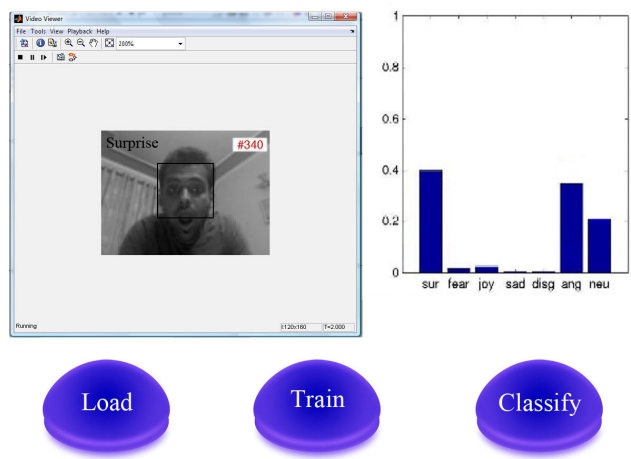

## V. CONCLUSION

Electronic commerce could also benefit from this technology. During the e-commerce buying process, the computer would be able to identify potential buyers' gestures, determine whether or not they intend to make a purchase and even gauge how satisfied they are with a product or service by helping to reduce the ambiguities of spoken or written language.

Another application is in security system. For instance, the technology could be used as a security measure at ATM's; instead of using a bank card or personal identification number, the ATM would capture an image of your face, and compare it to your photo in the bank database to confirm your identity. This same concept could also be applied to computers; by using a webcam to capture a digital image of yourself, your face could replace your password as a means to log-in.

Automatic expression recognition is a difficult task, which is afflicted by the usual difficulties faced in pattern recognition and computer vision research circles, coupled with face specific problems. However the correlation and Mahalanobis distance based approach explained in the paper is designed to accurately recognize the gestures in still as well as real time video sequence.

## ACKNOWLEDGEMENT

We are heartily thankful to Dr.T.V.Prasad(HOD C.S.E Dept, Lingaya's Institute Of Management & Technology) & Mr. Brijesh(HOD I.T Dept, Lingaya's Institute Of Management & Technology) whose encouragement, guidance and support from the initial to the final level enabled us to develop an understanding of the subject. We are deeply grateful to our Project Guides, Mr.Gautam Dutta (IT Dept.) & Mr.Bhanu Kiran(CSE Dept) , for their detailed and constructive comments, and for their important support throughout this work.

## AUTHORS PROFILE

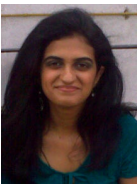
**Supriya Kapoor**, a final year computer science student at Lingaya's Institute of Mgt. & Tech., Faridabad, Haryana, India. Her areas of interest include Image processing, Artificial Neural Networks, software development life cycle, and project management-areas.

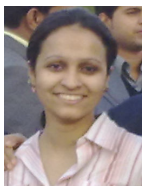
**Shruti Khanna**, a final year computer science student at Lingaya's Institute of Mgt. & Tech., Faridabad, Haryana, India. Her areas of interest include Image processing, Computer architecture and Artificial Neural-Networks.

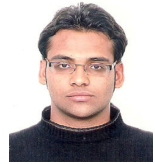
**Rahul Bhatia**, a final year information technology student at Lingaya's Institute of Mgt. & Tech., Faridabad, Haryana, India. His areas of interest include Image processing, Artificial Neural Networks, Computer organization and Operating System.